\documentclass[compsoc]{IEEEtran}
\usepackage{enumitem}
\usepackage{amssymb}
\usepackage{amsbsy}
\usepackage{amsmath}
\usepackage{amsthm} 
\usepackage{cleveref}
\usepackage{array}
\usepackage{float}
\usepackage{graphicx}
\usepackage{subcaption}
\usepackage{bm}
\usepackage[noadjust]{cite}

\usepackage{relsize}
\usepackage{stfloats}
\def\mathbi#1{\textbf{\em #1}}

\DeclareMathAlphabet\mathbfcal{OMS}{cmsy}{b}{n}
\usepackage[linesnumbered,ruled]{algorithm2e}

\ifCLASSOPTIONcompsoc
\else
\fi
\ifCLASSINFOpdf
\else
\fi
\begin{document}
\title{Online Multilinear Dictionary Learning}
%
%
%
%

\author{Thiernithi~Variddhisa\"i~and~Danilo~Mandic
        

\thanks{T. Variddhisa\"{i} and D. Mandic are with the Department
of Electrical \& Electronic Engineering, Imperial College London, United Kingdom.\protect\\
E-mails: \{tt1513,d.mandic\}@imperial.ac.uk
}
}

\IEEEcompsoctitleabstractindextext{%
\begin{abstract}
Tensors are a type of data structure used in signal processing to represent `naturally multidimensional' data. Its main benefit is the compact statistics e.g. covariance for `naturally multidimensional' data thanks to its Kronecker structure. There have also appealed to many applications of data fusion where the data which is not necessarily multidimensional can be fused with others to perform joint analysis for their hidden relationship (which is inherently assumed to be Kronecker). A method for online tensor dictionary learning is proposed. With the assumption of separable dictionaries, alternating stochastic gradients are used to diminish a $N$-way model of $\mathcal{O}\left(L^N\right)$ into a simple matrix equation of $\mathcal{O}\left(NL^2\right)$ with a real-time capability. This is prefered over the recursive least squares method because Kronecker product causes the inversed dimension to exponentially expand instead of unity as in standard RLS. Moreover, the joint tensor compressed sensing is also proposed along which accommodate online computation. Experiments on two synthetic signals and one real-world data verify the improved performance of the online method against the batch ones.
\end{abstract}

}

\maketitle

\IEEEdisplaynotcompsoctitleabstractindextext

%
\IEEEpeerreviewmaketitle

\section{Introduction}
%
%

%
%
%
%
\IEEEPARstart{W}{e} are witnessing a surge in the availability of data in many aspects of economy and life. This has posed unprecedented challenges, accompanied by opportunities, in gaining more profound insights into the information conveyed and hence enhanced understanding and better use of such data. Data acquired from today's large-scale sensors is usually complex, redundant and above all, massive in volume. This ``use of dimensionality'' is also a blessing if we can make sense of and later capitalize on the sheer volume and variety of data. In other words, this calls for sophisticated exploratory data analysis, as the direct brute-force treatment would be computationally and algorithmically prohibitive~\cite{Dhar2013}. The prerequisite when analyzing this deluge of data is the dimensionality reduction (a.k.a. sketching), that is, how to unveil the underlying compact representations (features) of data which in turn can be efficiently employed at a more sophisticated task e.g. inference and prediction. Furthermore, data with multiple explanatory features can also be viewed as multidimensional data where each dimension represents a feature. However, these features (a machine-learning term for variables) are largely unknown \textit{a priori} and thus need to be \textit{learned}. The transform of raw data into a set of its own features is known as representation learning~\cite{Bengio2013}.

Without priors on raw data, there exist an infinite number of possible representations. Although dependent on the data at hands, some of these priors can be general-purpose and ubiquitous in most types of data. The simplest assumed prior would be the orthogonality whereby all the features are mutually orthogonal. The orthogonality prior arises naturally when an analog signal is converted to a digital one via basic uniform sampling, leading to its digital representation as the sum of \textit{temporal} Delta functions (standard Euclidean bases in \textit{time} domain). Many classes of orthogonal bases have been proposed since, namely the Fourier bases (\textit{spectral} Delta functions) and wavelets~\cite{Daubechies1992}. It soon became apparent that this constraint is too restrictive for real-world data and the obtained features (bases), albeit with neat theoretical properties, are usually without practical meanings~\cite{Rubin2010}. One of alternative assumptions would be based on the fact that, for a specific physical phenomenon, there are a large number of possible (not necessarily orthogonal) features, but only few dominate at a time instance. This assumption is known as sparsity and collection of those features is termed \textit{representation dictionary}.

Dictionary learning (DL) is a class of feature learning grounded in the field of matrix factorization, and there is a considerable body of research in the literature~\cite{Aharon2006,Engan2007,Blumensath2009}, most of which are offline method and not suitable for the data which are either available sequentially or are too massive to analyze in a single batch. This problem can be addressed by the method of online dictionary learning (ODL) and the earliest one followed an LMS-inspired algorithm with rank-$1$ stochastic gradient~\cite{Aharon2008}. Another algorithm, based on block co-ordinate descent (BCD), utilizes past information as a means to form its cost function, giving improved performance~\cite{Mairal2010}. Several ensuing ODL works considered other aspects including recursive-least-square (RLS)-DL~\cite{Skretting2010}, discriminative learning~\cite{Zhang2012,Yang2014}, kernel dictionary~\cite{Kim2015} and ODL with pruning~\cite{Nader2016}.

Besides DL, compressed sensing (CS) is another fast-growing field which leverages the sparsity prior~\cite{Candes2006}. Work in CS demonstrates that, if \textit{sparse} or \textit{compressible} in some transformed bases, the data can be accurately represented by samples fewer than those from the Shannon-Nyquist criteria~\cite{Candes2008}, leading to data compression. Originally deploying random measurements~\cite{Candes2006}, the sensing schemes have developed into an optimization problem. One popular scheme is based on the closest tight-frame Gram matrix~\cite{Zelnik2011,Abolghasemi2012,Wassell2013}, with its many extensions such as the designs robust to measurement error~\cite{Li2015} and joint optimization of projection matrix and dictionary~\cite{Chen2013,Bai2015}.

Even with the synergy between ODL and CS, it will not suffice when the data is extraordinarily large, which highlights the limitations of standard flat-view matrix models. Here, tensors offer a more versatile and natural framework with richer theoretical attributes, such as rank, uniqueness etc. Like matrix factorization, tensor decomposition provides several efficient tools for handling massive multidimensional data~\cite{Cichocki2015}. For example, the Canonical Polyadic Decomposition (CPD) and Tucker Decomposition (TD) are generalizations of matrix SVD and PCA. The dictionary learning based on tensor modelling - the multilinear dictionary learning (MDL) - is therefore a logical step forward with many existing algorithms. Beginning with the concept of higher-order compressed sensing (HO-CS)~\cite{Duarte2012}, many matrix-based DL methods have been extended to tensors, namely the Kronecker OMP~\cite{Caiafa2013}, K-CPD~\cite{Duan2012}, K-HOSVD~\cite{Roemer2014}, T-MOD~\cite{Zhao2016} and the joint optimization between MDL and HO-CS~\cite{Ding2017}. To our best knowledge, no online algorithm of MDL has yet been proposed. To this end, our aim is to introduce the method of online multilinear dictionary learning (OMDL), which is inspired by adaptive filtering theory. Our two main contributions are:
\begin{itemize}[leftmargin=5mm]
\item[$\bullet$] Online tensor-based dictionary learning algorithm based on accelerated gradient methods;
\item[$\bullet$] Joint online design of the mode-wise projection matrices for sequential HO-CS via separable equiangular tight-frame (ETF) approximation of the target Gram matrix;
\end{itemize}

The remainder of this paper is organized as follows: Section 2 reviews relevant backgrounds in DL, CS and tensors. Section 3 is devoted to our proposed OMDL, and Section 4 to the online algorithm of mode-wise projection matrices. Section 5 presents a linear version of the corresponding algorithm as well as some extensions. Several experiments are studied in Section 6, and Section 7 concludes the paper.


\IEEEpubidadjcol

\section{Background Theories}
Given the relative maturity of the area, only important concepts will be summarized here.

\subsection{Products in Tensors}
For detailed reading, we refer to~\cite{Comon2014,Verv2014,Cichocki2015,Papa2017}. A tensor can be deemed a multi-way array whereby its `ways' or \textit{modes} are the order of the tensor; these can also be explanatory variables of the tensor data. A real-valued tensor of order $N$ is symbolized by a boldface calligraphic uppercase letter as $\mathbfcal{A},\mathbfcal{B}\in\mathbb{R}^{I_1\times I_2\cdots\times I_N}$ with its scalar entries by italic lowercase letters as $a_{i_1 i_2 \dots i_N}$. Conversely, a matrix, denoted by a boldface uppercase letter, $\mathbf{A}\in\mathbb{R}^{I_1\times I_2}$, can be considered a 2$^{nd}$-order tensor. Working with tensors requires various types of products to be defined. Similar to matrices, we can define the Frobenius inner product of two tensors $\mathbfcal{A},\mathbfcal{B}$ as
\begin{equation*}
{\left\langle \mathbfcal{A},\mathbfcal{B}\right\rangle}_F\triangleq\sum\limits_{i_1=1}^{I_1}\sum\limits_{i_2=1}^{I_2}\cdots\sum\limits_{i_N=1}^{I_N}a_{i_1 i_2 \dots i_N}b_{i_1 i_2 \dots i_N},
\end{equation*}
and as a special case, the Frobenius norm of a tensor $\mathbfcal{A}$ then will be $\left\Vert \mathbfcal{A} \right\Vert _{F}\triangleq\sqrt{{\left\langle \mathbfcal{A},\mathbfcal{A}\right\rangle}_F}$.

\noindent Given a matrix $\mathbf{B}^{(n)}\in\mathbb{R}^{J_n\times I_n}$, The mode-$n$ multilinear product between $\mathbfcal{A}$ and $\mathbf{B}^{(n)}$ yields another tensor given by
\vspace{-0.4cm}
\begin{equation*}
{\left(\mathbfcal{A}\times_n\mathbf{B}^{(n)}\right)}_{i_1 \dots i_{n-1} j_n i_{n+1} \dots i_N}\triangleq\sum\limits_{i_n=1}^{I_n}a_{i_1 i_2 \dots i_N}b_{j_n i_n}
\end{equation*}
where $\mathbf{B}$ is a mode-$n$ factor matrix. This gives rise to the more general full multilinear product
\begin{equation*}
[\![ \mathbfcal{A}\,;\mathbf{B}^{(1)},\mathbf{B}^{(2)},\dots,\mathbf{B}^{(N)} ]\!]\triangleq\mathbfcal{A}\times_1\mathbf{B}^{(1)}\times_2\mathbf{B}^{(2)}\cdots\times_N\mathbf{B}^{(N)}.
\end{equation*}
Note that there is no particular order of operation for each mode-$n$ product. Now given $\mathbfcal{C}\in\mathbb{R}^{J_1\times J_2\cdots\times J_M}$ and $I_n=J_m=K$, we can define the mode-$(n,m)$ contracted product (contraction) between $\mathbfcal{A}$ and $\mathbfcal{C}$ which yields an $(N+M-2)^{th}$-order tensor
\vspace{-0.2cm}
\begin{multline*}
{(\mathbfcal{A}{\times_m^n}\mathbfcal{C})}_{i_1 \dots i_{n-1} i_{n+1} \dots i_N j_1 \dots j_{m-1} j_{m+1} \dots j_M} \\
\triangleq\sum\limits_{k=1}^{K}a_{i_1 \dots i_{n-1} k i_{n+1} \dots i_N}c_{j_1 \dots j_{m-1} k j_{m+1} \dots j_M}.
\end{multline*}

Although the definition above displays contraction in a single common mode, tensors can be contracted in several modes or even in all modes. For tensors $\mathbfcal{A},\mathbfcal{B}$, we define the all-mode contraction as $\mathbfcal{A}{\times_{\mathbi{N}}^{\mathbi{N}}}\mathbfcal{B}$ where $\mathbi{N}=\{1,2,\dots,N\}$, which can be verified to equal ${\left\langle \mathbfcal{A},\mathbfcal{B}\right\rangle}_F$; this signifies that tensor contraction is able to fuse and reduce dimensionality of tensors. In fact, the essence of our proposed OMDL relies on the mode-wise operation where two tensors is \textit{contracted} into a matrix via the mode-`all-but-$n$' between $\mathbfcal{A}$ and $\bar{\mathbfcal{B}}$, for $\bar{\mathbfcal{B}}\in\mathbb{R}^{I_1\cdots I_{n-1}\times J_n\times I_{n+1}\cdots\times I_N}$, which diminishes to a matrix given by
\vspace{-0.2cm}
\begin{multline*}
{(\mathbfcal{A}{\times_{/n}^{/n}}\bar{\mathbfcal{B}})}_{i_n  j_n}= \\
\sum\limits_{i_1=1}^{I_1}\cdots\sum\limits_{i_{n-1}=1}^{I_{n-1}}\sum\limits_{i_{n+1}=1}^{I_{n+1}}\cdots\sum\limits_{i_N=1}^{I_N}a_{i_1 i_2 \dots i_N}\bar{b}_{i_1 \dots i_{n-1} j_n i_{n+1} \dots i_N}
\end{multline*}
where $/n$ denotes $\mathbi{N}/n=\{1,2,\dots,n-1,n+1,\dots,N\}$.

\subsection{Linear Dictionary Learning}
In the classical matrix setting, the goal of (linear) dictionary learning (DL) is to identify a representation dictionary, $\boldsymbol{\Psi}\in\mathbb{R}^{J \times L}$, which is overcomplete ($J<L$) and sparsely represents a signal of interest, $\mathbf{x}\in\mathbb{R}^J$, expressed in linear equations as $\mathbf{x}\cong\boldsymbol{\Psi}\mathbf{s}$ where $\mathbf{s}\in\mathbb{R}^L$ is the sparse representation of the target signal $\mathbf{x}$ over $\boldsymbol{\Psi}$-transform. Note that boldface lowercase letters indicate vectors. The sparse vector, $\mathbf{s}$, is $S$-sparse if it has only $S$ non-zero elements, i.e. $\left\Vert \mathbf{s} \right\Vert _0\leq S$. Since the dictionary is overcomplete (underdetermined), the sparsity prior is used as a constraint for a unique solution.

A classical problem considers a finite \textit{unlabeled} training set of $t$ signals of interest, $\mathbf{X}=[\mathbf{x}_1,\mathbf{x}_2,\dots,\mathbf{x}_t]\in\mathbb{R}^{J \times t}$, with their corresponding sparse representation, $\mathbf{S}=[\mathbf{s}_1,\mathbf{s}_2,\dots,\mathbf{s}_t]\in\mathbb{R}^{L \times t}$ over the dictionary $\boldsymbol{\Psi}$, and can be cast into:
\vspace{-0.5cm}
\begin{equation}
\label{eq:1}
\begin{gathered}
\underset{\boldsymbol{\Psi},\mathbf{S}}{\min} \ {\sum\limits_{\tau=1}^t}w_{\tau}\ell_u(\mathbf{x}_{\tau},\boldsymbol{\Psi},\mathbf{s}_{\tau}) \\
\textnormal{s.t.} \ \boldsymbol{\Psi}\in\mathbi{C}\subset\mathbb{R}^{J\times L}\, \textnormal{and} \, \left\Vert \mathbf{s}_{\tau} \right\Vert _0\leq S, \, \forall \tau\in\mathbi{t}
\end{gathered}
\end{equation}
where $w_{\tau}\geq0$ is a weighting parameter, $\mathbi{t}=\{1,2,\dots,t\}$, $\mathbi{C}$ is a constraint space of $\boldsymbol{\Psi}$, and $\ell_u(\cdot)$ is a loss function where the index $u$ emphasizes that the DL problem is \textit{unsupervised}. The most widely used loss function is in the linear least-squares form:
\vspace{-0.3cm}
\begin{equation}
\label{eq:2}
\ell_u(\mathbf{x}_{\tau},\boldsymbol{\Psi},\mathbf{s}_{\tau})={\left\Vert \mathbf{x}_{\tau}-\boldsymbol{\Psi}\mathbf{s}_{\tau} \right\Vert}_2^2.
\end{equation}
Since both $\boldsymbol{\Psi}$ and $\mathbf{S}$ are unknown, the optimization problem in~\eqref{eq:1} is non-convex. A popular approach to its solution is to use alternating minimization between $\boldsymbol{\Psi}$ and $\mathbf{S}$, known as the \textit{dictionary update} and the \textit{sparse coding}.

In the dictionary update step, let $\ell_u(\mathbf{x}_{\tau},\boldsymbol{\Psi})=\ell_u(\mathbf{x}_{\tau},\boldsymbol{\Psi},\mathbf{\hat{s}}_{\tau})$ where $\mathbf{\hat{s}}_{\tau}, \forall \tau\in\mathbi{t}$ is the optimal solution to the sparse coding problem in the preceding alternate step. The problem in~\eqref{eq:1} then changes accordingly to
\vspace{-0.2cm}
\begin{equation}
\label{eq:3}
\underset{\boldsymbol{\Psi}}{\min} \ {\sum\limits_{\tau=1}^t}w_{\tau}\ell_u(\mathbf{x}_{\tau},\boldsymbol{\Psi}) \ \ \textnormal{s.t.} \ \boldsymbol{\Psi}\in\mathbi{C}\subset\mathbb{R}^{J\times L}.
\end{equation}
\noindent The role of the constraint space $\mathbi{C}$ is to prevent $\boldsymbol{\Psi}$ from becoming arbitrarily large. Such constraint could be a unit column $\ell_2$-norm. A training pair $(\mathbf{x}_{\tau},\mathbf{\hat{s}}_{\tau})$ can then be utilized in either \textit{batch} (\cite{Aharon2006,Engan2007,Blumensath2009}) or \textit{online} (\cite{Aharon2008,Mairal2010,Skretting2010}) manner.

In the sparse coding step, let $\ell_u(\mathbf{x}_{\tau},\mathbf{s}_{\tau})=\ell_u(\mathbf{x}_{\tau},\boldsymbol{\hat{\Psi}},\mathbf{s}_{\tau})$ where $\boldsymbol{\hat{\Psi}}$ is the optimal value from the previous dictionary update step. Likewise,~\eqref{eq:1} now changes to
\vspace{-0.2cm}
\begin{equation*}
\underset{\mathbf{S}}{\textnormal{min}} \ {\sum\limits_{\tau=1}^t}w_{\tau}\ell_u(\mathbf{x}_{\tau},\mathbf{s}_{\tau}) \ \ \textnormal{s.t.} \ \left\Vert \mathbf{s}_{\tau} \right\Vert _0\leq S, \, \forall \tau\in\mathbi{t}.
\end{equation*}
Since each loss function depends on a single different $\mathbf{s}_{\tau}$, the problem above can be independently solved for each $\mathbf{s}_{\tau}$, as
\begin{equation}
\label{eq:4}
\underset{\mathbf{s}_{\tau}}{\textnormal{min}} \ \ell_u(\mathbf{x}_{\tau},\mathbf{s}_{\tau}) \ \ \textnormal{s.t.} \ \left\Vert \mathbf{s}_{\tau} \right\Vert _0\leq S, \, \forall \tau\in\mathbi{t}.
\end{equation}
Notice that sparse coding has existed since the fixed dictionaries (overcomplete Fourier and wavelets) long before the whole DL problem. Moreover, it also forms the crux of the reconstruction problem in compressed sensing.

\subsection{Compressed Sensing}
The compressed sensing (CS) paradigm~\cite{Candes2006,Candes2008,Qaisar2013} aims to unify data acquisition and compression through accurate recovery of the signal, $\mathbf{s}$, from a measurement signal, $\mathbf{y}\in\mathbb{R}^I$ with $I<L$. In the most basic sense, as $\mathbf{s}$ is sparse, it can be accurately recovered from $\mathbf{y}$ by solving~\cite{Candes2008}
\begin{equation*}
\underset{\mathbf{s}}{\textnormal{min}} \ \left\Vert \mathbf{s} \right\Vert _0 \ \ \textnormal{s.t.} \ \mathbf{y}=\boldsymbol{\Theta}\mathbf{s}.
\end{equation*}
where $\boldsymbol{\Theta}\in\mathbb{R}^{I \times L}$ is called a \textit{sensing matrix}. However, natural signals are rarely explicitly sparse; however, most do have sparse representation. Assuming that the signal of interest, $\mathbf{x}_{\tau}$, is in the form $\mathbf{x}_{\tau}=\boldsymbol{\Psi}\mathbf{s}_{\tau}, \, \forall \tau\in\mathbi{t}$, as described above, a full reconstruction problem~\cite{Qaisar2013} can be expressed as
\begin{equation}
\label{eq:5}
\underset{\mathbf{s}_{\tau}}{\min} \ \left\Vert \mathbf{s}_{\tau} \right\Vert _0 \ \ \textnormal{s.t.} \ \mathbf{y}_{\tau}=\boldsymbol{\Theta}\mathbf{s}_{\tau}\triangleq\boldsymbol{\Phi}\boldsymbol{\Psi}\mathbf{s}_{\tau}, \, \forall \tau\in\mathbi{t}
\end{equation}
where $\boldsymbol{\Phi}\in\mathbb{R}^{I \times J}$ is termed a \textit{projection matrix}. Note that~\eqref{eq:5} is merely an alternative statement of the sparse coding problem in~\eqref{eq:4}, with $\mathbf{y}_{\tau}$ substituted for $\mathbf{x}_{\tau}$, $\boldsymbol{\Theta}$ for $\boldsymbol{\Psi}$, and~\eqref{eq:2} for the loss function $\ell_u$.

Owing to the $\ell_0$-norm, the problem in~\eqref{eq:4} and~\eqref{eq:5} is NP-hard to solve exactly. Many sparse coding techniques exist in the literature, such as greedy algorithms, $\ell_1$ relaxation or Bregman iteration~\cite{Qaisar2013}, which can approximate the solution with arbitrarily small error under certain conditions, i.e. restricted isometry property (RIP) or mutual coherence~\cite{Candes2008}, on the sensing matrix $\boldsymbol{\Theta}$. This poses another challenge in CS, apart from sparse coding, to design the projection matrix $\boldsymbol{\Phi}$ so that $\boldsymbol{\Theta}$ satisfies those conditions\footnote{It was proved that when these conditions hold true, the signals of interest, $\mathbf{x}_{\tau}$, can be recovered from $\boldsymbol{\Phi}$ with $I<J$~\cite{Candes2011}.}.

To date, many different approaches have appeared and are mainly grounded on the mutual coherence of $\boldsymbol{\Theta}$, denoted by $\mu(\boldsymbol{\Theta})$. The problem boils down to designing the projection matrix $\boldsymbol{\Phi}$ such that the Gram matrix of $\boldsymbol{\Theta}$, defined as $\boldsymbol{\Theta}^T\boldsymbol{\Theta}$, is as close as possible to a target equiangular tight-frame (ETF) Gram matrix $\boldsymbol{\Gamma}\in\mathbi{G}_{\underline{\mu}}$ through the following optimization~\cite{Wassell2013}
\begin{equation}
\label{eq:6}
\underset{\boldsymbol{\Theta}}{\min} \ {\Vert \boldsymbol{\Gamma}-\boldsymbol{\Theta}^T\boldsymbol{\Theta} \Vert}^2_F\triangleq\underset{\boldsymbol{\Phi}}{\min} \ {\Vert \boldsymbol{\Gamma}-\boldsymbol{\Psi}^T\boldsymbol{\Phi}^T\boldsymbol{\Phi}\boldsymbol{\Psi} \Vert}^2_F
\end{equation}
where $\mathbi{G}_{\underline{\mu}}$ is a set of relaxed ETF Gram matrices defined as
\begin{equation}
\label{eq:7}
\begin{split}
\mathbi{G}_{\underline{\mu}}\triangleq\{\boldsymbol{\Gamma}\in\mathbb{R}^{L \times L}: \, &\boldsymbol{\Gamma}=\boldsymbol{\Gamma}^T, \, \textnormal{diag}(\boldsymbol{\Gamma})=1, \\
&\underset{i \neq j}{\textnormal{max}} \, |\boldsymbol{\Gamma}(i,j)|\leq\underline{\mu}\}.
\end{split}
\end{equation}
The parameter $\underline{\mu}$ is the lower bound of $\mu(\boldsymbol{\Theta})$ given by~\cite{Strohmer2003}
\begin{equation}
\label{eq:8}
\underline{\mu}=\sqrt{\frac{L-I}{I(L-1)}}\leq\mu(\boldsymbol{\Theta})\leq 1.
\end{equation}
Since the problem in~\eqref{eq:6} is highly non-convex, and likely to end up stuck in a local minimum, this is usually tackled by an iterative algorithm where the target $\boldsymbol{\Gamma}$ needs to be gradually updated so that the values of $\boldsymbol{\Theta}$ do not change too significantly~\cite{Zelnik2011,Abolghasemi2012,Wassell2013,Li2015}.
\section{Online Dictionary Learning for Tensors}
Many methods of block-based dictionary learning have been extended to tensors, and we have introduced a real-time online version, suitable for exceedingly large or steaming data. Here, a simplified accelerated first-order methods~\cite{Nesterov1983,Yim2017} are incorporated into a mode-wise coordinate descent method to derive an algorithm for tensor-based online learning of representation dictionaries. In this work, we use the Kronecker OMP, which guarantees the local minimum of the solution~\cite{Caiafa2013} for the sparse coding step.


\subsection{Multilinear Dictionary Learning - Preliminaries}
Let $\mathbfcal{X}^{(\tau)}\in\mathbb{R}^{J_1\times J_2\times\cdots\times J_N}, \, \forall \tau\in\mathbi{t}$ be an observed sequence of $t$ $N^{th}$-order tensors. Its multi-way sparse representation can be expressed in the form of a multilinear product~\cite{Duarte2012}
\begin{equation}
\label{eq:9}
\mathbfcal{X}^{(\tau)}=\mathbfcal{S}^{(\tau)}\times_1\boldsymbol{\Psi}_1\times_2\boldsymbol{\Psi}_2\cdots\times_N\boldsymbol{\Psi}_N+\mathbfcal{E}^{(\tau)}, \, \forall \tau\in\mathbi{t},
\end{equation}
where $\boldsymbol{\Psi}_n\in\mathbb{R}^{J_n\times L_n}$ is a mode-$n$ overcomplete dictionary (i.e. $J_n<L_n$), $\, \forall n\in\mathbi{N}$, $\mathbfcal{S}^{(\tau)}\in\mathbb{R}^{L_1\times L_2\times\cdots\times L_N}$ are sparse tensors associated with $\mathbfcal{X}^{(\tau)}$, and $\mathbfcal{E}^{(\tau)}$ are the error and noise. The tensor $\mathbfcal{S}^{(\tau)}$ is called $S$-sparse if it has only $S$ non-zero elements, that is, much fewer than the total dimension of the observation, i.e. $S\ll\prod_{n=1}^NJ_n$. Given fixed $S$-sparse tensors $\mathbfcal{S}^{(\tau)}, \, \forall \tau\in\mathbi{t}$, a multilinear extension of the dictionary update problem in~\eqref{eq:3} is given by
\vspace{-0.2cm}
\begin{equation}
\label{eq:10}
\underset{\left\{\boldsymbol{\Psi}\right\}}{\min} \ {\sum\limits_{\tau=1}^t}w^{(\tau)}\ell_u(\mathbfcal{X}^{(\tau)},\{\boldsymbol{\Psi}\}) \ \ \textnormal{s.t.} \ \boldsymbol{\Psi}_n\in\mathbi{C}_n, \, \forall n\in\mathbi{N}.
\end{equation}
where $\{\boldsymbol{\Psi}\}=\{ \boldsymbol{\Psi}_n,\, \ \forall n\in\mathbi{N}\}$ is a set of all mode-wise dictionaries, $\mathbi{C}_n\subset\mathbb{R}^{J_n\times L_n}$ is a mode-$n$ constraint space curbing the values of $\boldsymbol{\Psi}_n$. Now, the loss function $\ell_u(\cdot)$ in~\eqref{eq:2} will take a \textit{multilinear} least-square form~\cite{Roemer2014,Zhao2016}:
\vspace{-0.2cm}
\begin{multline}
\label{eq:11}
\ell_u(\mathbfcal{X}^{(\tau)},\{\boldsymbol{\Psi}\})= \\
{\Vert \mathbfcal{X}^{(\tau)}-\mathbfcal{S}^{(\tau)}\times_{1}\boldsymbol{\Psi}_1\times_{2}\boldsymbol{\Psi}_2\cdots\times_{N}\boldsymbol{\Psi}_N \Vert}_F^2.
\end{multline}

Even with fixed $\mathbfcal{S}^{(\tau)}$, solving~\eqref{eq:10} is a non-convex problem due to its multilinear structure. However, we can solve for each mode-$n$ dictionary by fixing the other modes based on rather natural condition that all mode-$n$ dictionaries are separable (i.e. each multilinear atom is only in the form of a rank-1 tensor, not of block-term one), which is known as the alternating linear scheme~\cite{Hawe2013}. Hence, let $\mathcal{J}^{(t)}(\{\boldsymbol{\Psi}\})$, or $\mathcal{J}^{(t)}$ in short, be an empirical objective function, built on~\eqref{eq:10} and~\eqref{eq:11} and defined in a mode-$n$ expression as
\begin{equation}
\label{eq:12}
\mathcal{J}^{(t)}\triangleq\frac{1}{2}{\sum\limits_{\tau=1}^t}w^{(\tau)}\left\Vert \mathbfcal{X}^{(\tau)}-\mathbfcal{\tilde{S}}^{(\tau)}_n\times_n\boldsymbol{\Psi}_n\right\Vert _{F}^{2},
\end{equation}
where $\tilde{\mathbfcal{S}}^{(\tau)}_n=\mathbfcal{S}^{(\tau)}\times_1\boldsymbol{\Psi}_1\times_2\boldsymbol{\Psi}_2\cdots\times_{n-1}\boldsymbol{\Psi}_{n-1}\times_{n+1}\boldsymbol{\Psi}_{n+1}\cdots\times_N\boldsymbol{\Psi}_N$. By utilizing the relationship between the matrix trace and Frobenius inner product, the right-hand side of~\eqref{eq:12} can be disentangled, with the help of contracted products, into a quadratic form of pure matrices, as in~\eqref{eq:13} shown at the bottom of the page, where the notation Tr$(\cdot)$ is the trace of a matrix.
\begin{figure*}[!b]
\normalsize
\hrulefill
\begin{equation}
\label{eq:13}
\small{\mathcal{J}^{(t)}=\textnormal{Tr}\left(\frac{1}{2}\boldsymbol{\Psi}_n\left({\sum\limits_{\tau=1}^t}w^{(\tau)}\left[\tilde{\mathbfcal{S}}^{(\tau)}_n\times_{/n}^{/n}\tilde{\mathbfcal{S}}^{(\tau)}_n\right]\right)\boldsymbol{\Psi}_n^T-\left({\sum\limits_{\tau=1}^t}w^{(\tau)}\left[\mathbfcal{X}^{(\tau)}\times_{/n}^{/n}\tilde{\mathbfcal{S}}^{(\tau)}_n\right]\right)\boldsymbol{\Psi}_n^T+\frac{1}{2}{\sum\limits_{\tau=1}^t}w^{(\tau)}\left[\mathbfcal{X}^{(\tau)}\times_{/n}^{/n}\mathbfcal{X}^{(\tau)}\right]\right)}
\end{equation}
\end{figure*}
With~\eqref{eq:13}, the all-mode optimization in~\eqref{eq:10} is `unfolded' into $n$ mode-wise problems as
\begin{equation}
\label{eq:14}
\underset{\boldsymbol{\Psi}_n}{\min} \ \mathcal{J}^{(t)} \ \ \textnormal{s.t.} \ \boldsymbol{\Psi}_n\in\mathbi{C}_n.
\end{equation}
Many recent works in MDL are based on the alternating least squares in~\eqref{eq:14} (\cite{Roemer2014,Zhao2016,Ding2017}), all of which consider the offline case where $t$ training pairs $(\mathbfcal{X}^{(\tau)},\mathbfcal{\tilde{S}}^{(\tau)}_n), \, \forall \tau\in\mathbi{t}$ are considered altogether at each iteration $n$.


\subsection{Alternating Linear Scheme for OMDL}
For an online implementation of~\eqref{eq:14}, the training pairs $(\mathbfcal{X}^{(\tau)},\mathbfcal{\tilde{S}}^{(\tau)}_n)$ are used one by one; in other words, the time instant $t$ grows progressively as a data pair is fed. For each mode-$n$ expression in~\eqref{eq:12}, let $w^{(\tau)}={\lambda}^{t-\tau}$ and $\tilde{\mathbf{S}}^{(\tau)}_n$ and $\tilde{\mathbf{Q}}^{(\tau)}_n$ be
\vspace{-0.2cm}
\begin{equation}
\label{eq:15}
\mathbf{S}^{(t)}_n\triangleq\tilde{\mathbfcal{S}}^{(t)}_n\times_{/n}^{/n}\tilde{\mathbfcal{S}}^{(t)}_n,
\end{equation}
\begin{equation}
\label{eq:16}
\mathbf{Q}^{(t)}_n\triangleq\mathbfcal{X}^{(t)}\times_{/n}^{/n}\tilde{\mathbfcal{S}}^{(t)}_n.
\end{equation}
Since the rightmost term of~\eqref{eq:13} does not depend on $\boldsymbol{\Psi}_n$,~\eqref{eq:14} is equivalent to
\vspace{-0.4cm}
\begin{equation}
\label{eq:17}
\underset{\boldsymbol{\Psi}_n}{\min} \ \mathcal{\hat{J}}^{(t)} \ \ \textnormal{s.t.} \ \boldsymbol{\Psi}_n\in\mathbi{C}_n
\end{equation}
where
\vspace{-0.2cm}
\begin{equation}
\label{eq:18}
\mathcal{\hat{J}}^{(t)}=\textnormal{Tr}\left(\frac{1}{2}\boldsymbol{\Psi}_n\mathbf{R}^{(t)}_n{\boldsymbol{\Psi}}_n^T-\mathbf{P}^{(t)}_n{\boldsymbol{\Psi}}_n^T\right)
\end{equation}
with the following recursive formulae:
\begin{equation}
\label{eq:19}
\mathbf{R}^{(t)}_n\triangleq{\sum\limits_{\tau=1}^t}{\lambda}^{t-\tau}\left[\tilde{\mathbfcal{S}}^{(\tau)}_n\times_{/n}^{/n}\tilde{\mathbfcal{S}}^{(\tau)}_n\right]=\lambda\mathbf{R}^{(t-1)}_n+\mathbf{S}^{(t)}_n,
\end{equation}
\begin{equation}
\label{eq:20}
\mathbf{P}^{(t)}_n\triangleq{\sum\limits_{\tau=1}^t}{\lambda}^{t-\tau}\left[\mathbfcal{X}^{(\tau)}\times_{/n}^{/n}\tilde{\mathbfcal{S}}^{(\tau)}_n\right]=\lambda\mathbf{P}^{(t-1)}_n+\mathbf{Q}^{(t)}_n,
\end{equation}
and $\lambda \in (0,1]$ is a forgetting parameter similar to that of an RLS algorithm.

To implement the mode-wise block coordinate descent method based on~\eqref{eq:18}, the gradient descent is extrapolated via stochastic conjugate direction~\cite{Yim2017}, whereby the descent direction takes the form
\vspace{-0.1cm}
\begin{equation}
\label{eq:21}
\mathbf{D}^{(t)}_n=-\mathbf{G}^{(t)}_n+\beta^{(t)}_n\mathbf{D}^{(t-1)}_n
\end{equation}
and the mode-$n$ gradient of $\mathcal{J}^{(t)}$ is given by
\vspace{-0.1cm}
\begin{equation}
\label{eq:22}
\mathbf{G}^{(t)}_n\triangleq\left.\frac{\partial\mathcal{J}^{(t)}}{\partial\boldsymbol{\Psi}_n}\right\vert_{\boldsymbol{\Psi}_n=\boldsymbol{\Psi}^{(t-1)}_n}=\boldsymbol{\Psi}^{(t-1)}_n\mathbf{R}^{(t)}_n-\mathbf{P}^{(t)}_n.
\end{equation}
The following theorem,
\newtheorem{mythm}{Theorem}
\begin{mythm}[\cite{Yim2017}]
A set of matrices $\{\mathbf{D}_n^{(1)},\mathbf{D}_n^{(2)},...,\mathbf{D}_n^{(t)}\}$ of the form~\eqref{eq:21} and satisfying
\begin{equation*}
\textnormal{Tr}\left(\mathbf{D}_n^{(t-1)}\mathbf{R}_n^{(t)}\mathbf{D}_n^{(t)^T}\right)=0, \ \forall t,
\end{equation*}
is a descent direction of the objective function~\eqref{eq:18},
\end{mythm}
\noindent allows us to obtain $\beta^{(t)}_n$ as
\begin{equation}
\label{eq:23}
\beta^{(t)}_n=\frac{{\left\langle\mathbf{H}^{(t)}_n,\mathbf{G}^{(t)}_n\right\rangle}_F}{{\left\langle\mathbf{H}^{(t)}_n,\mathbf{D}^{(t-1)}_n\right\rangle}_F}
\end{equation}
where $\langle\cdot,\cdot\rangle_F$ is a Frobenius inner product and
\vspace{-0.1cm}
\begin{equation}
\label{eq:24}
\mathbf{H}^{(t)}_n=\mathbf{D}^{(t-1)}_n\mathbf{R}^{(t)}_n.
\end{equation}
Finally, the mode-$n$ dictionary is iteratively calculated as
\vspace{-0.1cm}
\begin{equation}
\label{eq:25}
\boldsymbol{\Psi}^{(t)}_n=\mathsf{\Pi}_{\mathbi{C}_n}\left[\boldsymbol{\Upsilon}^{(t)}_n\right]=\mathsf{\Pi}_{\mathbi{C}_n}\left[\boldsymbol{\Psi}^{(t-1)}_n+\mathbf{D}^{(t)}_n\mathbf{A}^{(t)}_n\right]
\end{equation}
where $\mathbf{A}^{(t)}_n$ is a diagonal matrix, of which the diagonals are $\alpha_n^{(t)}(l)=\mathbf{R}^{(t)}_n[l,l]$, and $\mathsf{\Pi}_{\mathbi{C}_n}\left[\cdot\right]$ is an orthogonal projector onto the convex set $\mathbi{C}_n$. For example, when the convex set $\mathbi{C}_n$ is a linear map which conserves a space spanned by the dictionary atoms i.e. the column space, this turns~\eqref{eq:25} into
\begin{equation}
\label{eq:26}
\boldsymbol{\Psi}^{(t)}_n=\boldsymbol{\Upsilon}^{(t)}_n\boldsymbol{\Pi}^{(t)}_n
\end{equation}
and $\boldsymbol{\Pi}^{(t)}_n$ is a diagonal matrix the diagonals of which, $\pi_n^{(t)}(l)$, are given by
\begin{equation}
\label{eq:27}
\pi_n^{(t)}(l)=\frac{1}{\textnormal{max}\left({\|\mathbf{u}_n^{(t)}(l)\|}_2,1\right)}, \, \forall l=1,2,\dots,L_n
\end{equation}
where $\mathbf{u}_n^{(t)}(l)$ is the $l^{th}$ column vector of $\boldsymbol{\Upsilon}^{(t)}_n$. The proposed tensor dictionary learning algorithm, named the online multilinear dictionary learning (OMDL), is summarized in Algorithm 1, where $\delta>0$ is used as a stopping criterion.
\begin{algorithm}
\SetKwInOut{Input}{Input}
\SetKwInOut{Output}{Output}
\SetKwRepeat{DoWhile}{do}{while}
\Input{$\mathbfcal{X}^{(t)}\in\mathbb{R}^{J_1\times J_2\times\cdot\cdot\times J_N}$ (inputs), $T$ (number of inputs), $\boldsymbol{\Psi}_n^{(0)}\in\mathbb{R}^{J_n \times L_n}$ (initial dictionaries), $N$ (number of modes), $\lambda$ (forgetting factor)}
\Output{$\boldsymbol{\Psi}_n^{(t)}$ (modewise dictionaries)}
 Initialize $\mathbf{R}^{(0)}_n=\mathbf{\underline{0}}$, $\mathbf{P}^{(0)}_n=\mathbf{\underline{0}}$ and $\mathbf{D}^{(0)}_n=\mathbf{\underline{0}}$ \ $\forall n$\;
 \For{t = 1 \textnormal{to} T}{
  Obtain sparse core tensor $\mathbfcal{S}^{(t)}$ via appropriate sparse coding scheme e.g.~\cite{Caiafa2013}\;
  \For{n = 1 \textnormal{to} N}{
  Update $\mathbf{S}^{(t)}_n$ and $\mathbf{Q}^{(t)}_n$ by~\cref{eq:15,eq:16}\;
  $\mathbf{R}^{(t)}_n=\lambda\mathbf{R}^{(t-1)}_n+\mathbf{S}^{(t)}_n$\;
  $\mathbf{P}^{(t)}_n=\lambda\mathbf{P}^{(t-1)}_n+\mathbf{Q}^{(t)}_n$\;
  
  $\mathbf{G}^{(t)}_n=\boldsymbol{\Psi}^{(t-1)}_n\mathbf{R}^{(t)}_n-\mathbf{P}^{(t)}_n$\;
  $\mathbf{H}^{(t)}_n=\mathbf{D}^{(t-1)}_n\mathbf{R}^{(t)}_n$\;
  $\beta^{(t)}_n=\frac{{\left\langle\mathbf{H}^{(t)}_n,\mathbf{G}^{(t)}_n\right\rangle}_F}{{\left\langle\mathbf{H}^{(t)}_n,\mathbf{D}^{(t-1)}_n\right\rangle}_F},$ \ ($\beta^{(1)}_n=0$)\;
  $\mathbf{D}^{(t)}_n=-\mathbf{G}^{(t)}_n+\beta^{(t)}_n\mathbf{D}^{(t-1)}_n$\;
  
  Update $\mathbf{A}^{(t)}_n$ where its diagonals $\alpha_n^{(t)}(l)=\mathbf{R}^{(t)}_n[l,l]$\;
  $\boldsymbol{\Upsilon}^{(t)}_n=\boldsymbol{\Psi}^{(t-1)}_n+\mathbf{D}^{(t)}_n\mathbf{A}^{(t)}_n$\;
  Update $\boldsymbol{\Pi}^{(t)}_n$ by~\cref{eq:27}\;
  $\boldsymbol{\Psi}^{(t)}_n=\boldsymbol{\Upsilon}^{(t)}_n\boldsymbol{\Pi}^{(t)}_n$\;
  }
 }
\caption{The OMDL Algorithm}
\end{algorithm}

\subsection{Insights into Convergence}
\begin{figure}[H]
\centering
\includegraphics[width=\columnwidth]{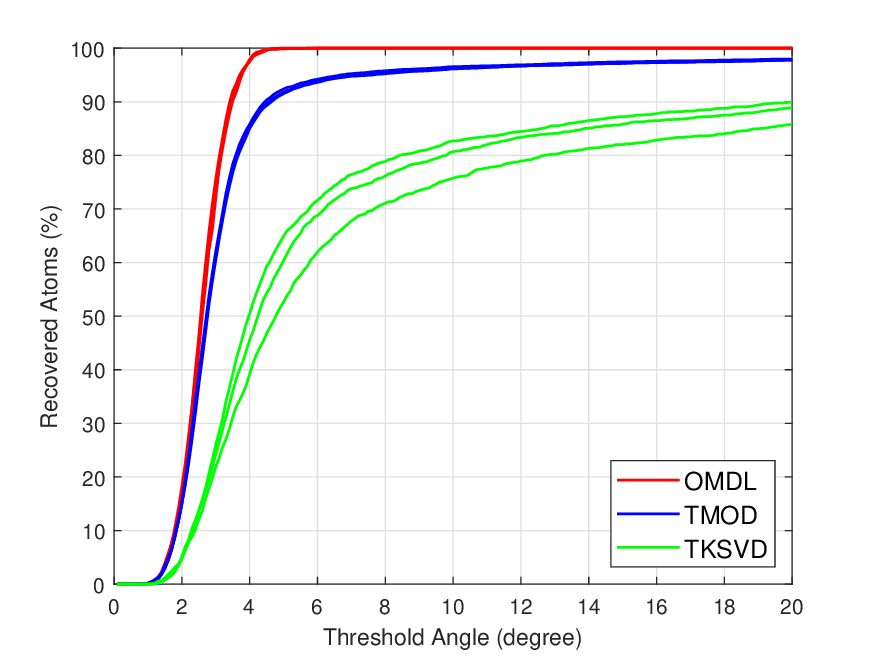}
\caption{Successful recovery of atoms with respect to different 'threshold' angle for all 3 modes grouped in the same color, with red (MODL), blue (TMOD) and green (TKSVD) ($L_n=20$, $J_n=10$, $S_n=8$, $n=1,2,3$)}
\label{fig:1}
\end{figure}

The sparse coding stage typically governs the overall convergence analysis~\cite{Bengio2013,Rubin2010} because the \textit{online} dictionary update stage is a variant of Least Mean Square algorithm, a form of quadratic programming with well-understood convergence property~\cite{Mandic2015}. When sparse coding gives the global optimal solution, e.g. through LASSO~\cite{Keni2017}, then so does the whole algorithm. Here, the experiments were performed in a synthetic scenario where the sparse core is assumed known to show how accurately the dictionary atoms are recovered. The core tensor $\mathbfcal{S}^{(t)}\in\mathbb{R}^{20\times 20\times 20}$ with equal mode-wise sparsity, $S_n=8 (n=1,2,3)$, has non-zero elements randomly selected from a Gaussian distribution and each mode-$n$ dictionary, $\boldsymbol{\Psi}_n\in\mathbb{R}^{10\times 20} (n=1,2,3)$, is generated by Gaussian random variable with mean and variance equal $0$ and $1$ respectively. With SNR set to $0$ dB, the $3^{rd}$-order tensor data, $\mathbfcal{X}^{(t)}\in\mathbb{R}^{10\times 10\times 10}$, was generated via~\eqref{eq:9}. The success of recovery of the mode-wise dictionary $\boldsymbol{\Psi}_n\in\mathbb{R}^{10\times 20} (n=1,2,3)$ was measured by $\theta$, the angle between $\boldsymbol{\psi}_{real}$ and $\boldsymbol{\psi}_{learned}$, the real and the recovered atoms (vectors) respectively. If the angle is below some \textit{threshold}, the atom is successfully recovered i.e.
\vspace{-0.1cm}
\begin{equation*}
\frac{\boldsymbol{\psi}_{real}\cdot\boldsymbol{\psi}_{learned}}{|\boldsymbol{\psi}_{real}||\boldsymbol{\psi}_{learned}|}>cos(\theta).
\end{equation*}
This test was run over 100 trials to compare three similar tensor algorithms: the proposed MODL, TMOD~\cite{Zhao2016} and the TKSVD~\cite{Ding2017} as shown in Fig.~\ref{fig:1}.

From Fig.~\ref{fig:1}, the MODL algorithm was able to recover all atoms within roughly $5$ threshold degrees while the other algorithms considered were unable to do so, even with more relaxed threshold of 20 degrees. For a far insight, the TKSVD performed worse possibly because this experiment assumes known sparse core while the TKSVD updates the core consistently, even off the sparse coding step. Overall, this result illuminates that, given an effective sparse coding scheme, the alternating stochastic gradient of the MODL algorithm does not impede global convergence \textit{per se}. A rigorous proof of convergence to stationary point is given in~\cite{Mairal2010} for standard dictionary learning, and we will not give the OMDL counterpart as it would trivially follow the same mechanism.
\vspace{-0.4cm}
\section{Joint Design for Sequential HO-CS}
So far, compressed sensing has been extended into tensor models and even incorporated into the multilinear dictionary learning. However, for tensors, this joint design attempt has been made only for the offline case and with a target Gram matrix for each mode is the identity~\cite{Ding2017}. Also, it is quite common in many related works to include the effects of projected error to the design of the optimal projection matrices~\cite{Li2015,Bai2015}. In our work, a more relaxed ETF scheme is employed through a simplified robust design. This simplification is possible due to the observation that the size of the projected error from the projection matrices is dictated by the sizes of the sparse representation error (SRE) and the projection matrix under some practical conditions. In case of tensors, this notion can be straightforwardly applied. 

\subsection{Higher-Order Compressed Sensing - Preliminaries}
The higher-order compressed sensing (HO-CS) is a multilinear extension of the CS problem in~\eqref{eq:5}. Building upon~\eqref{eq:9}, the HO-CS task can be written as
\vspace{-0.2cm}
\begin{equation}
\label{eq:31}
\begin{gathered}
\underset{\mathbfcal{S}^{(\tau)}}{\min} \ \left\Vert \mathbfcal{S}^{(\tau)} \right\Vert _0 \ \ \textnormal{s.t.} \\ \mathbfcal{Y}^{(\tau)}=\mathbfcal{S}^{(\tau)}\times_1\boldsymbol{\Theta}_1\times_2\boldsymbol{\Theta}_2\cdots\times_N\boldsymbol{\Theta}_N, \, \forall \tau\in\mathbi{t},
\end{gathered}
\end{equation}
where $\mathbfcal{Y}^{(\tau)}\in\mathbb{R}^{I_1\times I_2\cdots\times I_N}$ is measurement tensor signal and $\boldsymbol{\Theta}_N\triangleq\boldsymbol{\Phi}_N\boldsymbol{\Psi}_N\in\mathbb{R}^{I_n \times L_n}$, $\, \forall n\in\mathbi{N}$ is a mode-$n$ sensing matrix. With~\eqref{eq:9}, $\mathbfcal{Y}^{(\tau)}$ can be expressed in terms of $\mathbfcal{X}^{(\tau)}$ as
\begin{equation}
\label{eq:32}
\mathbfcal{Y}^{(\tau)}=\mathbfcal{X}^{(\tau)}\times_1\boldsymbol{\Phi}_1\times_2\boldsymbol{\Phi}_2\cdots\times_N\boldsymbol{\Phi}_N, \, \forall \tau\in\mathbi{t}
\end{equation}
where $\boldsymbol{\Phi}_n\in\mathbb{R}^{I_n \times J_n}$ is called a mode-$n$ projection matrix with $I_n\leq J_n$, $\, \forall n\in\mathbi{N}$. It is noteworthy that the problem in~\eqref{eq:31} is equivalent to the conventional CS problem with a Kronecker structure~\cite{Duarte2012}:
\begin{equation}
\label{eq:33}
\bar{\mathbf{y}}_{\tau}=\boldsymbol{\bar{\Theta}}\bar{\mathbf{s}}_{\tau}\triangleq\left(\boldsymbol{\Theta}_N\otimes\boldsymbol{\Theta}_{N-1}\otimes\cdots\otimes \ \boldsymbol{\Theta}_1\right)\bar{\mathbf{s}}_{\tau}
\end{equation}
where it is clear that, if we define $\boldsymbol{\bar{\Phi}}\triangleq\boldsymbol{\Phi}_N\otimes\boldsymbol{\Phi}_{N-1}\otimes\cdots\otimes \ \boldsymbol{\Phi}_1$ and $\boldsymbol{\bar{\Psi}}\triangleq\boldsymbol{\Psi}_N\otimes\boldsymbol{\Psi}_{N-1}\otimes\cdots\otimes \ \boldsymbol{\Psi}_1$ and use the mixed-product property of the Kronecker product, we have
\vspace{-0.1cm}
\begin{equation}
\label{eq:34}
\boldsymbol{\bar{\Theta}}=\boldsymbol{\bar{\Phi}}\boldsymbol{\bar{\Psi}}
\end{equation}
Based on the conventional Gram matrix problem in~\eqref{eq:6}-\eqref{eq:8}, we obtain the following problem:
\vspace{-0.1cm}
\begin{equation}
\label{eq:35}
\underset{\boldsymbol{\bar{\Theta}}}{\min} \ {\Vert \boldsymbol{\Gamma}-\boldsymbol{\bar{\Theta}}^T\boldsymbol{\bar{\Theta}} \Vert}^2_F=\underset{\boldsymbol{\bar{\Phi}}}{\min} \ {\Vert \boldsymbol{\Gamma}-\boldsymbol{\bar{\Psi}}^T\boldsymbol{\bar{\Phi}}^T\boldsymbol{\bar{\Phi}}\boldsymbol{\bar{\Psi}} \Vert}^2_F
\end{equation}
with $\boldsymbol{\Gamma}\in\mathbi{G}_{\underline{\mu}}$ given by
\begin{equation}
\label{eq:36}
\begin{split}
\mathbi{G}_{\underline{\mu}}\triangleq\{&\boldsymbol{\Gamma}\in\mathbb{R}^{L_1L_2\dots L_N \times L_1L_2\dots L_N}: \, \boldsymbol{\Gamma}={\boldsymbol{\Gamma}}^T, \\
&\textnormal{diag}(\boldsymbol{\Gamma})=1, \, \underset{i \neq j}{\textnormal{max}} \, |\boldsymbol{\Gamma}(i,j)|\leq\underline{\mu}\}.
\end{split}
\end{equation}
and
\vspace{-0.4cm}
\begin{equation}
\label{eq:37}
\underline{\mu}=\sqrt{\left.\left({\prod\limits_{n=1}^N}L_n-{\prod\limits_{n=1}^N}I_n\right)\middle/\left({\prod\limits_{n=1}^N}I_n({\prod\limits_{n=1}^N}L_n-1)\right)\right.}.
\end{equation}

While it is possible to use this conventional CS approach to solve~\eqref{eq:35}, the explicit manipulation of $\boldsymbol{\bar{\Theta}}$ can however be highly prohibitive owing to its very large dimension. Moreover, it is rather difficult to enforce the Kronecker structure into $\boldsymbol{\bar{\Theta}}$. It is shown that, via the separable structure of~\eqref{eq:31}, we can solve for individual mode-$n$ projection matrices, $\boldsymbol{\Phi}_n$, mode by mode~\cite{Caiafa2013,Rivenson2009} as long as each mode-$n$ projection matrix conforms to standard RIP and mutual coherence conditions (more rigorous theories can be found in~\cite{Duarte2012}). However, those approaches used an identity matrix as a target Gram matrix which inherently has Kronecker structure, while for $\boldsymbol{\Gamma}$ in~\eqref{eq:35}, this is not necessarily the case.

In order to solve~\eqref{eq:35} alternately, we instead solve a similar problem:
\begin{equation}
\vspace{-0.1cm}
\label{eq:38}
\underset{\boldsymbol{\bar{\Theta}}}{\min} \ {\Vert \boldsymbol{\bar{\Gamma}}-\boldsymbol{\bar{\Theta}}^T\boldsymbol{\bar{\Theta}} \Vert}^2_F=\underset{\boldsymbol{\bar{\Phi}}}{\min} \ {\Vert \boldsymbol{\bar{\Gamma}}-\boldsymbol{\bar{\Psi}}^T\boldsymbol{\bar{\Phi}}^T\boldsymbol{\bar{\Phi}}\boldsymbol{\bar{\Psi}} \Vert}^2_F
\end{equation}
where $\boldsymbol{\bar{\Gamma}}\triangleq\boldsymbol{\Gamma}_N\otimes\boldsymbol{\Gamma}_{N-1}\otimes\cdots\otimes \ \boldsymbol{\Gamma}_1$ with $\boldsymbol{\Gamma}_n\in\mathbi{G}_{\underline{\mu}_n}$ given by
\begin{equation}
\label{eq:39}
\begin{split}
\mathbi{G}_{\underline{\mu}_n}&\triangleq\{\boldsymbol{\Gamma}_n\in\mathbb{R}^{L_n \times L_n}: \, \boldsymbol{\Gamma}_n={\boldsymbol{\Gamma}}_n^T, \\
&\textnormal{diag}(\boldsymbol{\Gamma}_n)=1, \, \underset{i \neq j}{\textnormal{max}} \, |\boldsymbol{\Gamma}_n(i,j)|\leq\underline{\mu}_n\}.
\end{split}
\end{equation}
and
\vspace{-0.4cm}
\begin{equation}
\label{eq:40}
\underline{\mu}_n=\textnormal{min}\left(\sqrt{\frac{L_n-I_n}{I_n(L_n-1)}},\underline{\mu}\right).
\end{equation}
Here through~\eqref{eq:39} and~\eqref{eq:40}, $\boldsymbol{\bar{\Gamma}}$ is guaranteed to satisfy~\eqref{eq:36}, i.e. $\boldsymbol{\bar{\Gamma}}\in\mathbi{G}_{\underline{\mu}}$. This ensures both valid global solution and a separable Kronecker constraint for the mode-wise Gram matrices.

\subsection{Alternating Scheme for Mode-Wise Projection Matrix Design}
In order to design robust projection matrices, the projected error should be as small as possible for the corresponding CS system to perform well in practice~\cite{Abolghasemi2012,Li2015}. This is equal to adding the projected error as a regularizer into~\eqref{eq:35}, thus yielding the following optimization problem:
\begin{equation}
\label{eq:41}
\underset{\boldsymbol{\bar{\Phi}},\boldsymbol{\bar{\Gamma}}}{\min} \ {\Vert \boldsymbol{\bar{\Gamma}}-\boldsymbol{\bar{\Psi}}^T\boldsymbol{\bar{\Phi}}^T\boldsymbol{\bar{\Phi}}\boldsymbol{\bar{\Psi}} \Vert}^2_F+\sigma{\Vert\boldsymbol{\bar{\Phi}}\mathbf{e} \Vert}^2_F
\end{equation}
where $\mathbf{e}$ is the vectorized SRE, $\textnormal{vec}(\mathbfcal{E})$, defined in~\eqref{eq:9} and $\sigma$ is a weighting parameter. Without any assumptions, it is obvious that
\vspace{-0.1cm}
\begin{equation}
\label{eq:42}
{\Vert\boldsymbol{\bar{\Phi}}\mathbf{e}\Vert}_F \leq {\Vert\boldsymbol{\bar{\Phi}}\Vert}_F{\Vert\mathbf{e}\Vert}_2=\left({\prod\limits_{n=1}^N}{\Vert{\boldsymbol{\Phi}}_n\Vert}_F\right){\Vert\mathbfcal{E}\Vert}_F.
\end{equation}
In other words, the size of the projected error is bounded above by the sizes of the SRE and the projection matrices in all modes. Since ${\Vert\mathbfcal{E}\Vert}_F$ is minimized at the dictionary learning stage, then ${\prod_{n=1}^N}{\Vert{\boldsymbol{\Phi}}_n\Vert}_F$ can be considered a surrogate of ${\Vert\boldsymbol{\bar{\Phi}}\mathbf{e}\Vert}_F$ and minimized instead. Furthermore, if $\mathbfcal{E}$ can be modelled as Gaussian noise and the number of training data, $t$, is large enough, then the equality holds in~\eqref{eq:41}~\cite{Hong2018}. These assumptions therefore simplify~\eqref{eq:41} to
$$\underset{\boldsymbol{\bar{\Phi}},\boldsymbol{\bar{\Gamma}}}{\min} \ \mathcal{V}^{(t)}(\boldsymbol{\bar{\Phi}},\boldsymbol{\bar{\Gamma}})$$
\vspace{-0.2cm}
with
\begin{equation}
\label{eq:43}
\mathcal{V}^{(t)}(\boldsymbol{\bar{\Phi}},\boldsymbol{\bar{\Gamma}})={\Vert \boldsymbol{\bar{\Gamma}}-\boldsymbol{\bar{\Psi}}^{(t)^T}\boldsymbol{\bar{\Phi}}^T\boldsymbol{\bar{\Phi}}\boldsymbol{\bar{\Psi}}^{(t)} \Vert}^2_F+\sigma{\prod\limits_{n=1}^N}{\Vert{\boldsymbol{\Phi}}_n\Vert}^2_F
\end{equation}
where the optimal $\boldsymbol{\bar{\Phi}}$ is independent of $\mathbfcal{E}$. This significantly facilitates online computation.

To address this non-convex problem, an alternating minimization algorithm is used. It is worth noting that this expression is the same as the eq. (23) in~\cite{Ding2017} where alternating gradient descent is used for the non-separable approach employed in our paper as it was shown to outperform and more computationally efficient than the separable one. Firstly, a shrinking operation is applied to~\eqref{eq:43} to obtain $\boldsymbol{\bar{\Gamma}}$ mode by mode~\cite{Abolghasemi2012,Yaghoobi2009}. By defining $\boldsymbol{\Theta}_n^{(t)}\triangleq\boldsymbol{\Phi}_n^{(t)}\boldsymbol{\Psi}_n^{(t)}$ and $\boldsymbol{\Theta}_n^{(t^*)}\triangleq\boldsymbol{\Phi}_n^{(t-1)}\boldsymbol{\Psi}_n^{(t)}$, we now obtain
\begin{equation}
\label{eq:44}
\boldsymbol{\Gamma}_n^{(t)}[i,j]=\left\{\begin{array}{ll}
    \gamma_n(i,j), & |\gamma_n(i,j)|\leq\underline{\mu}_n \\
    \textnormal{sgn}[\gamma_n(i,j)]\underline{\mu}_n, & |\gamma_n(i,j)|>\underline{\mu}_n \\
    1, & i=j 
\end{array}\right.
\end{equation}
\begin{equation}
\label{eq:45}
\boldsymbol{\Gamma}_n^{(t^*)}[i,j]=\left\{\begin{array}{ll}
    \gamma_n^*(i,j), & |\gamma_n^*(i,j)|\leq\underline{\mu}_n \\
    \textnormal{sgn}[\gamma_n^*(i,j)]\underline{\mu}_n, & |\gamma_n^*(i,j)|>\underline{\mu}_n \\
    1, & i=j 
\end{array}\right.
\end{equation}
where $\gamma_n$ and $\gamma_n^*$ are the $(i,j)$-elements of the corresponding normalized Gram of the matrices $\boldsymbol{\Theta}_n^{(t)^T}\boldsymbol{\Theta}_n^{(t)}$ and $\boldsymbol{\Theta}_n^{(t^*)^T}\boldsymbol{\Theta}_n^{(t^*)}$ respectively. Then, $\boldsymbol{\bar{\Phi}}$ is iteratively calculated per mode. By defining the three following parameters:
\begin{equation}
\label{eq:46}
\rho_n^{(t)}={\prod\limits_{k=1}^{n-1}}{\Vert \boldsymbol{\Theta}_k^{(t)^T}\boldsymbol{\Theta}_k^{(t)} \Vert}^2_F{\prod\limits_{k=n+1}^N}{\Vert \boldsymbol{\Theta}_k^{(t^*)^T}\boldsymbol{\Theta}_k^{(t^*)} \Vert}^2_F
\end{equation}
\begin{equation}
\label{eq:47}
\begin{split}
\omega_n^{(t)}&={\prod\limits_{k=1}^{n-1}}\textnormal{Tr}\left( \boldsymbol{\Theta}_k^{(t)}\boldsymbol{\Gamma}_k^{(t)}\boldsymbol{\Theta}_k^{(t)^T}\right)\times \\
&{\prod\limits_{k=n+1}^N}\textnormal{Tr}\left( \boldsymbol{\Theta}_k^{(t^*)}\boldsymbol{\Gamma}_k^{(t^*)}\boldsymbol{\Theta}_k^{(t^*)^T} \right)
\end{split}
\end{equation}
\begin{equation}
\label{eq:48}
\zeta_n^{(t)}={\prod\limits_{k=1}^{n-1}}{\Vert \boldsymbol{\Phi}_k^{(t)} \Vert}^2_F{\prod\limits_{k=n+1}^N}{\Vert \boldsymbol{\Phi}_k^{(t-1)} \Vert}^2_F
\end{equation}
the update equation for projection matrix becomes
\begin{equation}
\label{eq:49}
\boldsymbol{\Phi}_n^{(t)}=\boldsymbol{\Phi}_n^{(t-1)}-\eta_n\mathbf{V}_n^{(t)}
\end{equation}
where $\eta_n$ is a stepsize parameter and the mode-wise gradient $\mathbf{V}_n^{(t)}$ is given in~\eqref{eq:50} below; all constants are absorbed into the regularizer $\sigma$. The joint optimization algorithm is summarized in Algorithm 2.
\begin{algorithm}
\SetKwInOut{Input}{Input}
\SetKwInOut{Output}{Output}
\SetKwRepeat{DoWhile}{do}{while}

\Input{$\mathbfcal{X}^{(t)}\in\mathbb{R}^{J_1\times J_2\times\cdot\cdot\times J_N}$ (inputs), $T$ (number of inputs), $\boldsymbol{\Psi}_n^{(0)}\in\mathbb{R}^{J_n \times L_n}$ (initial dictionaries), $\boldsymbol{\Phi}_n^{(0)}\in\mathbb{R}^{I_n \times J_n}$ (initial sensing matrices), $N$ (number of modes), $\lambda$ (forgetting factor)}
\Output{$\boldsymbol{\Phi}_n^{(t)}$ (modewise sensing matrices), $\boldsymbol{\Psi}_n^{(t)}$ (modewise dictionaries)}
 Initialize $\mathbf{R}^{(0)}_n=\mathbf{\underline{0}}$, $\mathbf{P}^{(0)}_n=\mathbf{\underline{0}}$ and $\mathbf{D}^{(0)}_n=\mathbf{\underline{0}}$ \ $\forall n$\;
 \For{t = 1 \textnormal{to} T}{
  Obtain $\boldsymbol{\Psi}_n^{(t)}$ $\forall n$ via the OMDL in Algorithm 1\;
  \For{n = 1 \textnormal{to} N}{
  \For{k = 1 \textnormal{to} n-1}{
  $\boldsymbol{\Theta}_k^{(t)}=\boldsymbol{\Phi}_k^{(t)}\boldsymbol{\Psi}_k^{(t)}$\;
  }
  \For{k = n \textnormal{to} N}{
  $\boldsymbol{\Theta}_k^{(t^*)}=\boldsymbol{\Phi}_k^{(t-1)}\boldsymbol{\Psi}_k^{(t)}$\;
  }
  $\gamma_n(i,j)=\boldsymbol{\Theta}_n^{(t)^T}\boldsymbol{\Theta}_n^{(t)}[i,j]$\;
  $\gamma_n^*(i,j)=\boldsymbol{\Theta}_n^{(t^*)^T}\boldsymbol{\Theta}_n^{(t^*)}[i,j]$\;
  
  Update $\boldsymbol{\Gamma}_n^{(t)}$ and $\boldsymbol{\Gamma}_n^{(t^*)}$ via~\cref{eq:44,eq:45}\;

  Calculate $\rho^{(t)}_n$, $\omega^{(t)}_n$, $\zeta^{(t)}_n$ via~\cref{eq:46,eq:47,eq:48}\;
  Obtain $\mathbf{V}_n^{(t)}$ via~\cref{eq:49}\;
  $\boldsymbol{\Phi}_n^{(t)}=\boldsymbol{\Phi}_n^{(t-1)}-\eta_n\mathbf{V}_n^{(t)}$\;
  }
 }
\caption{Joint Optimization Algorithm}
\end{algorithm}

\begin{figure*}[!b]
\normalsize
\hrulefill
\begin{equation}
\label{eq:50}
\mathbf{V}_n^{(t)}\triangleq\left.\frac{\partial\mathcal{V}(\boldsymbol{\bar{\Phi}},\boldsymbol{\bar{\Gamma}})}{\partial\boldsymbol{\Phi}_n}\right\vert_{\boldsymbol{\Phi}_n=\boldsymbol{\Phi}_n^{(t-1)}}=\rho_n^{(t)}\left[\boldsymbol{\Theta}_n^{(t^*)}\boldsymbol{\Theta}_n^{(t^*)^T}\boldsymbol{\Theta}_n^{(t^*)}\boldsymbol{\Psi}_n^{(t)^T}\right]-\omega_n^{(t)}\left[\boldsymbol{\Theta}_n^{(t^*)}\boldsymbol{\Gamma}_n^{(t^*)}\boldsymbol{\Psi}_n^{(t)^T}\right]+\sigma\zeta_n^{(t)}\left[\boldsymbol{\Phi}_n^{(t-1)}\right].
\end{equation}
\end{figure*}

\section{Experimental Validation}
A series of experiments were conducted to explore the performance of the proposed algorithms. The performance was evaluated against two criteria, the Normalized Root Mean Squared Error (NRMSE) and the Average Representation Error (ARE)~\cite{Chen2013}, respectively given by
\begin{equation}
\label{eq:51}
\sigma_{nrmse}=\frac{{\Vert\mathbfcal{X}-\mathbfcal{S}\times_1\boldsymbol{\Psi}_1\times_2\boldsymbol{\Psi}_2\cdots\times_N\boldsymbol{\Psi}_N\Vert}_F}{{\Vert\mathbfcal{X}\Vert}_F}
\end{equation}
\begin{equation}
\label{eq:52}
\sigma_{are}=\frac{{\Vert\mathbfcal{X}_{w/o}-\mathbfcal{S}\times_1\boldsymbol{\Psi}_1\times_2\boldsymbol{\Psi}_2\cdots\times_N\boldsymbol{\Psi}_N\Vert}_F}{\textnormal{lens}(\mathbfcal{X}_{w/o})}
\end{equation}
where lens($\cdot$) denotes the total number of elements of the operand.  

\vspace{-0.2cm}
\begin{figure}[H]
    \centering
    \begin{subfigure}[ht]{\columnwidth}
        \includegraphics[width=\columnwidth]{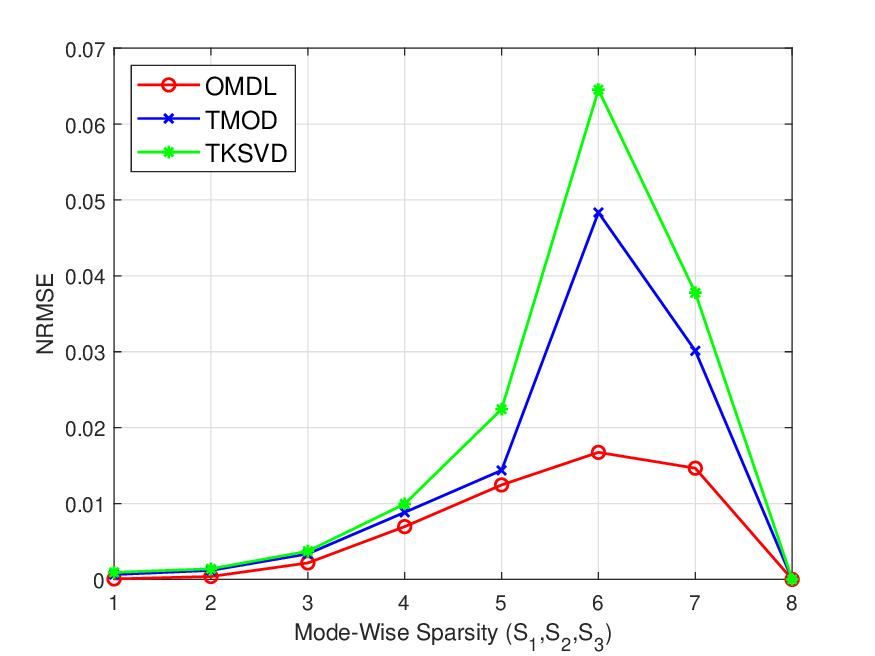}
        \caption{}
        \label{fig:2a}
    \end{subfigure}
    \begin{subfigure}[ht]{\columnwidth}
        \includegraphics[width=\columnwidth]{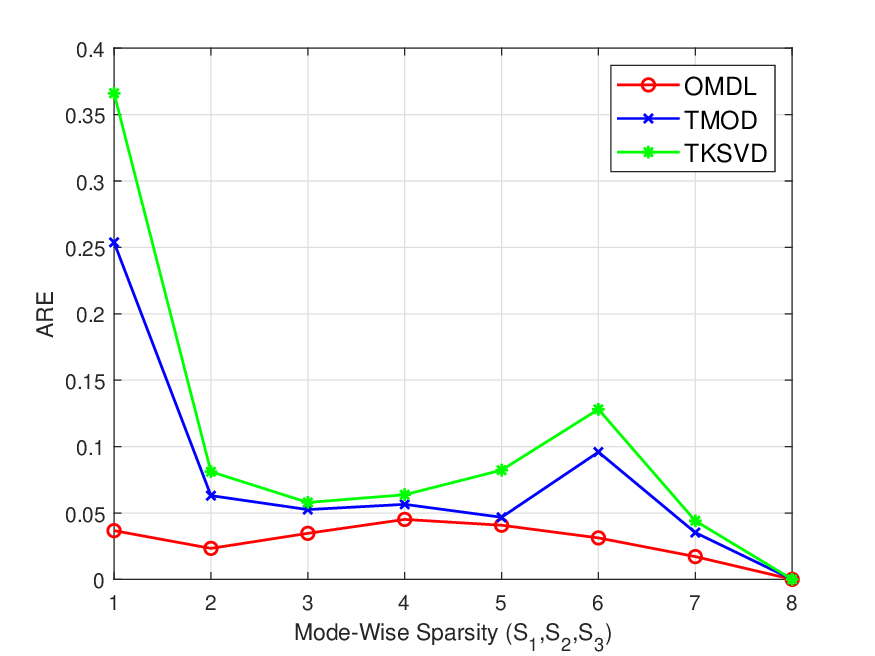}
        \caption{}
        \label{fig:2b}
    \end{subfigure}
    ~ 
    \caption{Performance measures, (a) NRMSE and (b) ARE, of three considered algorithms with respect to different sparsity levels, for 1000 3-mode tensor data $\mathbfcal{X}\in\mathbb{R}^{8\times 8\times 8}$ generated by full multilinear product between 1000 sparse core tensors $\mathbfcal{S}\in\mathbb{R}^{16\times 16\times 16}$ and mode-$n$ dictionary $\boldsymbol{\Psi}_n\in\mathbb{R}^{8\times 16},n=1,2,3$, with an average SNR of 20 dB and a forgetting factor $\lambda=0.95$, averaged over 100 realizations}
    \label{fig:2}
\end{figure}

\noindent\textit{A. Online Multilinear Dictionary Learning}

\vspace{0.2cm}
In the first experiment, we compared the dictionary learning stage of the OMDL with those of other similar algorithms, namely TMOD and TKSVD. A set of observed data contained 1000 3-mode tensors $\mathbfcal{X}\in\mathbb{R}^{8\times 8\times 8}$ generated by a full multilinear product between 1000 sparse core tensors $\mathbfcal{S}\in\mathbb{R}^{16\times 16\times 16}$ and mode-$n$ dictionary $\boldsymbol{\Psi}_n\in\mathbb{R}^{8\times 16},n=1,2,3$ with an average SNR of 20 dB and a forgetting factor $\lambda=0.95$. For the TMOD and TKSVD, we adjusted the hyperparameters according to~\cite{Zhao2016,Ding2017} so that they yielded the best respective performance for fair comparison; all simulations were averaged over 100 realizations.

\begin{figure}[ht]
    \centering
    \begin{subfigure}[ht]{\columnwidth}
        \includegraphics[width=\columnwidth]{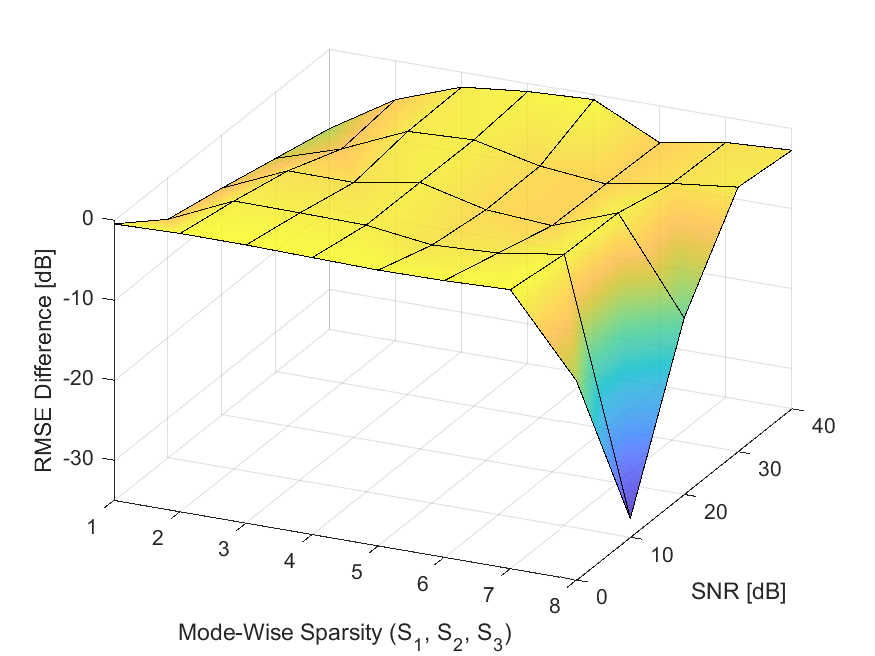}
        \caption{}
        \label{fig:3a}
    \end{subfigure}
    \begin{subfigure}[ht]{\columnwidth}
        \includegraphics[width=\columnwidth]{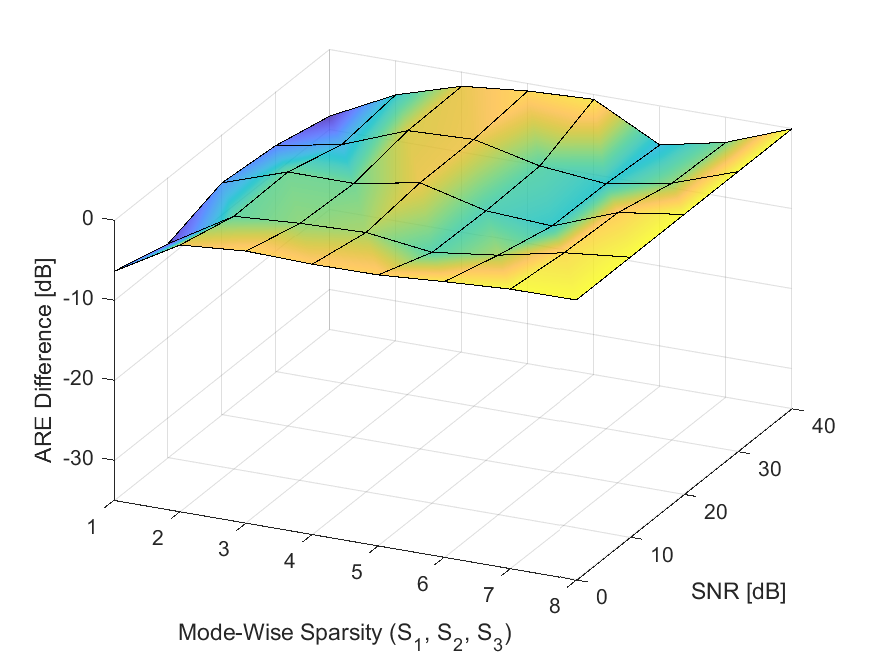}
        \caption{}
        \label{fig:3b}
    \end{subfigure}
    ~ 
    \caption{Performance difference, (a) NRMSE and (b) ARE, between OMDL and TMOD with respect to different sparsity and SNR, for 1000 3-mode tensor data $\mathbfcal{X}\in\mathbb{R}^{8\times 8\times 8}$ generated by a full multilinear product between 1000 sparse core tensors $\mathbfcal{S}\in\mathbb{R}^{16\times 16\times 16}$ and mode-$n$ dictionary $\boldsymbol{\Psi}_n\in\mathbb{R}^{8\times 16},n=1,2,3$, with the forgetting factor $\lambda=0.95$, averaged over 100 realizations}
    \label{fig:3}
\end{figure}

In Fig.~\ref{fig:2}, the two measures (NRMSE and ARE) were computed against different levels of sparsity of the core tensors for the three considered algorithms. For simplicity, the sparsity levels were  equal mode-wise sparsity i.e. block sparsity. As defined, the proposed online algorithm consistently outperformed the batch methods. Fig.~\ref{fig:3} compares the measure difference between the OMDL and TMOD with respect to different block sparsity and SNR. The surface has values below 0 dB, thus verifying the OMDL consistently yielded better performance than the TMOD.
\vspace{-0.2cm}
\vspace{-0.2cm}
\begin{figure}[ht]
    \centering
    \begin{subfigure}[ht]{\columnwidth}
        \includegraphics[width=\columnwidth]{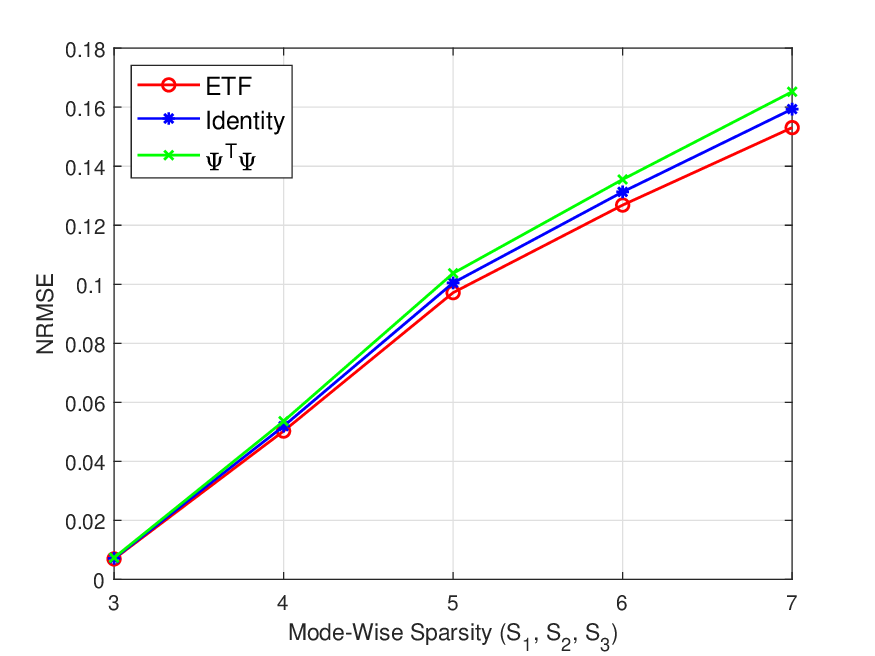}
        \caption{}
        \label{fig:4a}
    \end{subfigure}
    \begin{subfigure}[ht]{\columnwidth}
        \includegraphics[width=\columnwidth]{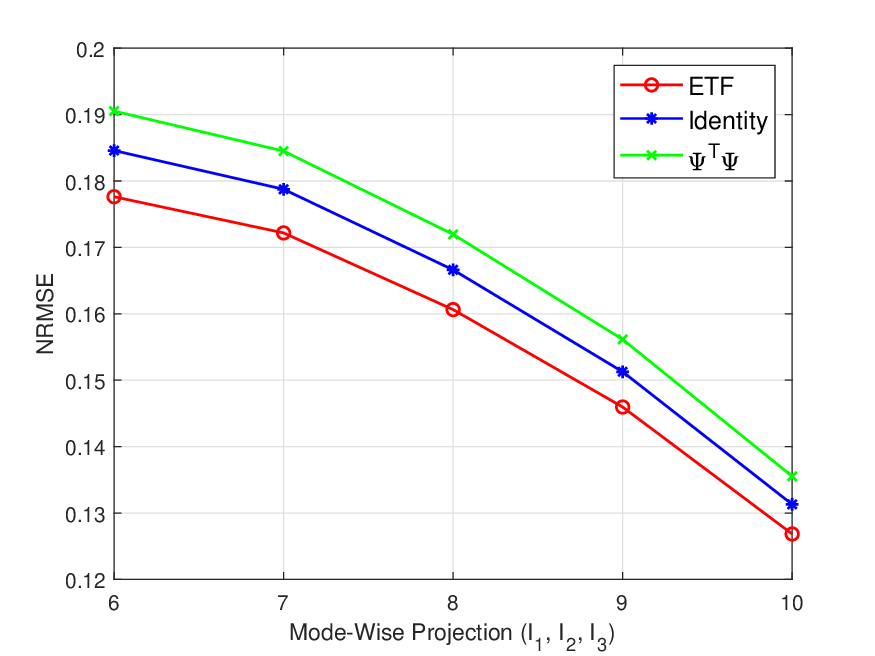}
        \caption{}
        \label{fig:4b}
    \end{subfigure}
    ~ 
    \caption{NRMSE Performance measures of different $\boldsymbol{\bar{\Gamma}}$ with respect to (a) different sparsity for $I_n=10,\forall n$, and (b) different projection size for $S_n=6,\forall n$, for 1000 3-mode tensor data $\mathbfcal{X}\in\mathbb{R}^{16\times 16\times 16}$ generated by a full multilinear product between 1000 sparse core tensors $\mathbfcal{S}\in\mathbb{R}^{40\times 40\times 40}$ and mode-$n$ dictionary $\boldsymbol{\Psi}_n\in\mathbb{R}^{16\times 40},n=1,2,3$, with an average SNR of 20 dB and a forgetting factor $\lambda=0.95$, averaged over 100 realizations}
    \label{fig:4}
\end{figure}

\vspace{-0.3cm}
\noindent\textit{B. Online Joint Learning of Multilinear Dictionary-Projection Matrices}
\vspace{0.2cm}

In the second experiment, we compared the performance of the whole joint optimization of the online method was assessed with different rules for $\boldsymbol{\bar{\Gamma}}$ for the proposed OMDL used. We generated 1000 3-mode tensor data $\mathbfcal{X}\in\mathbb{R}^{16\times 16\times 16}$ by a full multilinear product between 1000 sparse core tensors $\mathbfcal{S}\in\mathbb{R}^{40\times 40\times 40}$, and mode-$n$ dictionary, $\boldsymbol{\Psi}_n\in\mathbb{R}^{16\times 40},n=1,2,3$. With an average SNR of 20 dB and a forgetting factor $\lambda=0.95$ for all simulations, we tested for different values of block sparsity ($S_n$) and projection size ($I_n$), strictly $S_n\leq I_n$. The measure used in this experiment was the NRMSE of the reconstructed tensor data $\mathbfcal{X}$. We employed the multipath matching pursuit~\cite{Kwon2014} to recover the sparse core tensor $\mathbfcal{S}$.

With $\boldsymbol{\bar{\Gamma}}$ set as the proposed ETF scheme, identity and $\boldsymbol{\bar{\Psi}}^T\boldsymbol{\bar{\Psi}}$, respectively, Fig.~\ref{fig:4a} shows that when the data is extremely sparse ($S_n$ is small), there is no much difference in the way how $\boldsymbol{\bar{\Gamma}}$ is set. However, as the sparsity level increases, $\boldsymbol{\bar{\Gamma}}$ begins to affect the performance of the algorithm, with the ETF scheme starting to beat the others. In Fig.~\ref{fig:4b} where sparsity level was fixed ($S_n=6,n=1,2,3$), the ETF scheme clearly yielded better results. Observe that as the projection size grows, the more accurate the reconstruction becomes.

\vspace{-0.2cm}
\begin{figure}[ht]
    \centering
    \begin{subfigure}[ht]{\columnwidth}
        \includegraphics[width=\columnwidth]{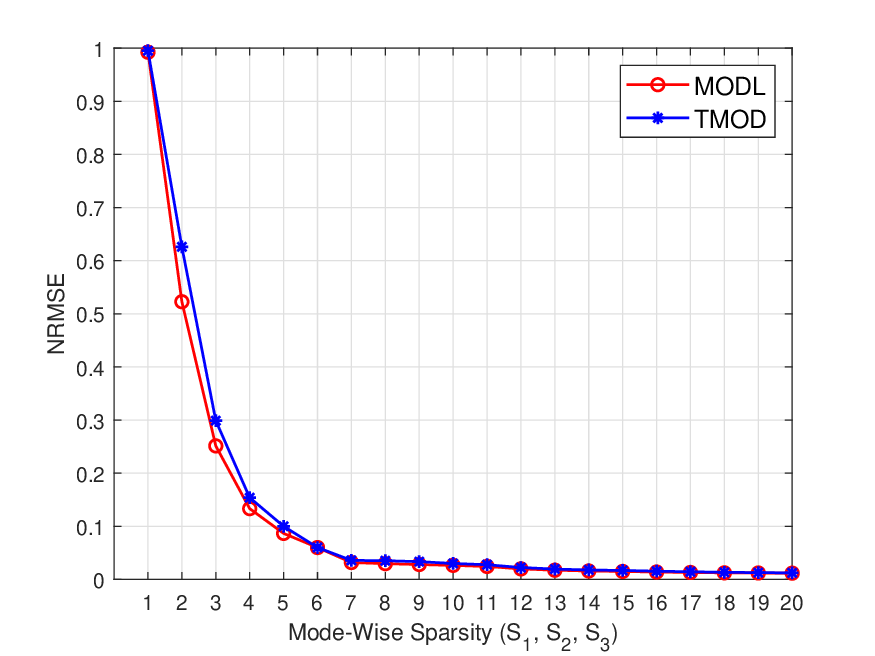}
        \caption{}
        \label{fig:5a}
    \end{subfigure}
    \begin{subfigure}[ht]{\columnwidth}
        \includegraphics[width=\columnwidth]{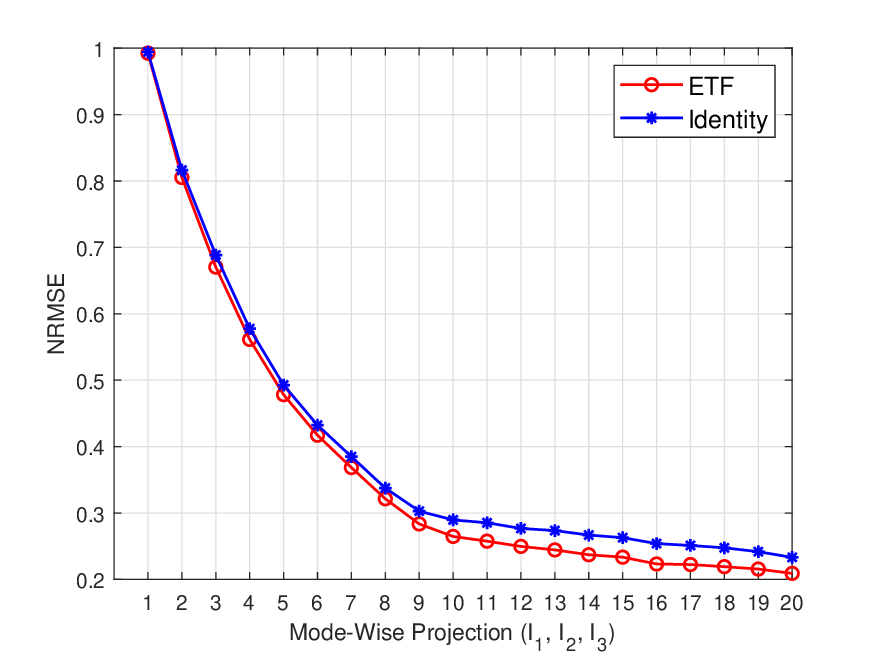}
        \caption{}
        \label{fig:5b}
    \end{subfigure}
    ~ 
    \caption{NRMSE Performance for (a) different dictionary learning schemes with varying sparsity and fixed $L_n=40,\forall n$, and (b) different compressed sensing schemes with varying projection size but fixed $J_n=15$ and $S_n=40,\forall n$, for 600 3-mode HSI patches of size ${24\times 24\times 24}$ extracted from the Indian Pines dataset of size $145\times145\times224$}
    \label{fig:5}
\end{figure}

\vspace{0.2cm}
\noindent\textit{C. Learning Performance on Hyperspectral Images}
\vspace{0.2cm}

Finally, we verify the performance of the proposed OMDL algorithm against real-world hyperspectral images sourced from the Indian Pins data of $145\times145$ pixels and $224$ spectral bands, i.e. a tensor of size $145\times145\times224$, collected by 1992 AVIRIS sensor. We randomly selected 600 patches each of size $24\times24\times24$ from the Indian Pines data for learning. First, the dictionary learning step was tested alone with the overcomplete sparse core tensor set at $40\times40\times40$ (overcompleteness level per mode $\approx1.6$). The NRMSE measure was used to compare the performance of MODL vs. TMOD algorithms with varying levels of mode-wise sparsity, ranging from $1$ to $20$. The second experiment tested joint optimization with MODL fixed as dictionary learning, but comparing the ETF and Identity scheme for the compressed sensing step. Here, the sparse core tensor is the same as previously and the mode-wise sparsity was set at $15$. The NRMSE was measured from the reconstructed results from different mode-wise projection size. The depth-first variant of multipath matching pursuit~\cite{Kwon2014} was applied on the vectorized data to recover the original HSI.

It can be seen from Fig.~\ref{fig:5a} that the MODL performed on par with its offline version. With an addition of peripheral variables like forgetting factor, stepsize etc, we could fine-tune the OMDL to yield marginally better results, but at no significant additional cost in terms of implementation. Fig.~\ref{fig:5b} confirms that the ETF scheme could, albeit slightly, reduce the reconstruction error compared to other compressed sensing strategies.
\section{Conclusion}
We have extended the jointly optimized dictionary-projection matrix learning for tensor data to the online learning paradigm. In this way, the batch TMOD method has been modified to operate in a sequential fashion, to obtain the online multilinear dictionary learning (OMDL) algorithm. In addition, we have also proposed modified compressed sensing for a Tucker model (HO-CS), where the target Gram matrix is relaxed from identity to equiangular tight frame (ETF). Although the overall performance improvement is not massive, it has been shown to enable a more computational-friendly method in cases where all the data may not be available altogether or computing all data at once is prohibitive. The advantages have been demonstrated by experiments on both synthetic and real-world data.


%

\ifCLASSOPTIONcaptionsoff
  \newpage
\fi


\renewcommand{\baselinestretch}{1}


\bibliographystyle{ieeetr}
\bibliography{test}


%









\end{document}